# High Accurate Unhealthy Leaf Detection

S. Mohan Sai[1], G. Gopichand[2], C. Vikas Reddy[3], K. Mona Teja[4]

[1,3,4] Under Graduate, Computer Science and Engineering, Vellore Institute of Technology, Vellore.

[2] Professor, Computer Science and Engineering, Vellore Institute of Technology, Vellore.

[1]singamsettymohansai10@gmail.com, [2]gopichand.g@vit.ac.in, [4]monateja9999@gmail.com

## Abstract:

India is an agriculture-dependent country. As we all know that farming is the backbone of our country it is our responsibility to preserve the crops. However, we can't stop the destruction of crops by natural calamities at least we have to try to protect our crops from diseases. To, detect a plant disease we need a fast automatic way. So, this paper presents a model to identify particular disease of plant leaves at early stages so that we can prevent or take a remedy to stop spreading of the disease. This proposed model is made into five sessions. Image preprocessing includes the enhancement of the low light image done using inception modules in CNN. Low-resolution image enhancement is done using an Adversarial Neural Network. This also includes Conversion of RGB Image to YCrCb color space. Next, this paper presents a methodology for image segmentation which is an important aspect for identifying the disease symptoms. This segmentation is done using the genetic algorithm. Due to this process the segmentation of the leaf Image this helps in detection of the leaf mage automatically and classifying. Texture extraction is done using the statistical model called GLCM and finally, the classification of the diseases is done using the SVM using Different Kernels with the high accuracy.

## I. Introduction:

The agriculture is the main of occupation of the many countries and cultures. Now a days, the land that is given for agriculture is only used for feeding source on planets earth. The economy of India primarily depends only on Agriculture Yield. Therefore, discovery of disease in the ground of farming is very important. The detection of the plants in the early stages is very important inorder to prevent the crops from spoiling completely. Different parts of the earth causes different diseases mainly in the United States there is a disease which is called as the little leaf. This is very harmful disease and it mainly occurs in the Douglas fir trees. This tree that is being affected with the disease will stop its development and perishes within six to seven years of lifespan. Its effect is discovered in many other parts of the world Georgian Sothern

America and Alabama. And some of the existing methods for the disease recognition in plants is by bare judgment seeing it with eyes by specialists over the color, texture where discovering and recognition of disease is done in different plants. For this, there is requirement of the huge crew of specialists and constant care taking of various plants is mandatory and it is very costly once if we deal through the huge farms. And some places of country doesn't have proper consultancy of meeting experts. In such cases this proposed model will be very highly used for the monitoring. Automatic detection of disease just by observing the leaf of the plant will be much cheaper. And the disease recognition that is done manually is very difficult job and also it may lead to low precise and can be completed in restricted place. Whereas if it is models are done mechanically, it leads in less effort and less time.

Image segmentation is a process that mainly emphases on the procedure of combining an image into various parts. The segmentation can be done in the different methods one basic approach is by the Threshold value. And then segmentation using colors. The computers have no sense of detecting of objects or separating so, many methods have been developed for the segmentation. This segmentation method proposed is by using the genetic algorithm. Various advantages of the genetic procedure is that this procedure improves very efficiently. It can improve high price exteriors. This paper is organized in five sections. The Section 1 is the introductory of the problem statement and the challenges in this current area. Section 2 Describes the Literature Survey Which is the previous working models of the unhealthy leaf detection. And the section 3 Describes about the architecture of the Model and the detail working of each step. Section 4 describes about the final outcomes of the research and the accurateness. And the last section 5 gives the conclusion and the Future work details.

## II. Literature review:

In this paper, the plant leaf disease can be identified using 4 main steps. The first step includes the taking of input that is the given image as the RGB image and later a colour that gets transformed is taken. It also uses some unique threshold value and the green colored pixels are marked, masked and detached using image segmentation techniques. At the final stage, the classifier is used in order to extract and also to classify the disease. [1] According to this paper, the plant leaf disease can be identified in the early and accurately using ANN i.e. Artificial Neural Network and many different digital image processing methodologies. The proposed methodology is mainly founded on the Artificial Neural Networks classifier aimed at the feature extraction, it gives an accurate percentage of finding the disease at 91%. [2] This paper tells that by using Principal Component Analysis (PCA) they can improve the performance of the leaf identification system. This method involves the features such as veins, colour, shape and the texture of the leaf of a plant. From different results, they suggested that PCA has the ability to increase the accuracy of the disease identification on the leaf. [3] According to this paper, the non-probabilistic algorithm using pixel-based is used for detecting the unhealthy regions that are the diseases that are caused on the leaves. This task is performed in three steps. First, segmentation is the process that is used in order to divide the image into frontend and the background. Support Vector Machines (SVM) are applied to expect the class of each pixel that is present on the front end of the leaf. The results are noted down and the proposed algorithm is designed for overcoming the plant disease. [4] In this paper, Content-Based Image Retrieval (CBIR) is used for getting visual features like color, shape, texture and these are used for the extraction of the given image. Other method is the Image Feature Extraction that takes place for the unhealthy leaf of a plant. Also, in Biometric systems, the images are used as input patterns such as iris, fingerprint, hand etc. [5] According to this paper, there are five steps used for finding the unhealthy plant. It includes Image Acquisition where we take a leaf of a plant and capture the digital image of it. Next step includes Image pre-processing where enhancement of the image is taken by means of SF-CES and also the color space conversion takes place in this step. Next step is Image Segmentation using K- Means clustering and the next step is extraction of features using GLCM. Last step is classification of the disease. [6] In this paper, it is mentioned about the automatic detection of diseases that are caused by fungal in wheat plants. In this to reduce the dimensionality of different features like size, color, texture, important features are designated by minimal redundancy maximal relevance criterion (mRMR). To classify the wheat diseases, A Radial Basis Function (RBF) is used. By using this method, we can find an accurate of 98.3% of wheat diseases. [7] In this paper, it presents us the technique of machine learning for recognizing the diseases in leaves. This is convenient technique because this saves much work, time in addition it also saves money. Method that used is co-occurrence method. The proposed approach can accurately detect the diseases caused on leaves and roots of a plants. [8]

# III. Proposed Methodology:

The structure of the proposed methodology is shown in Figure.1.

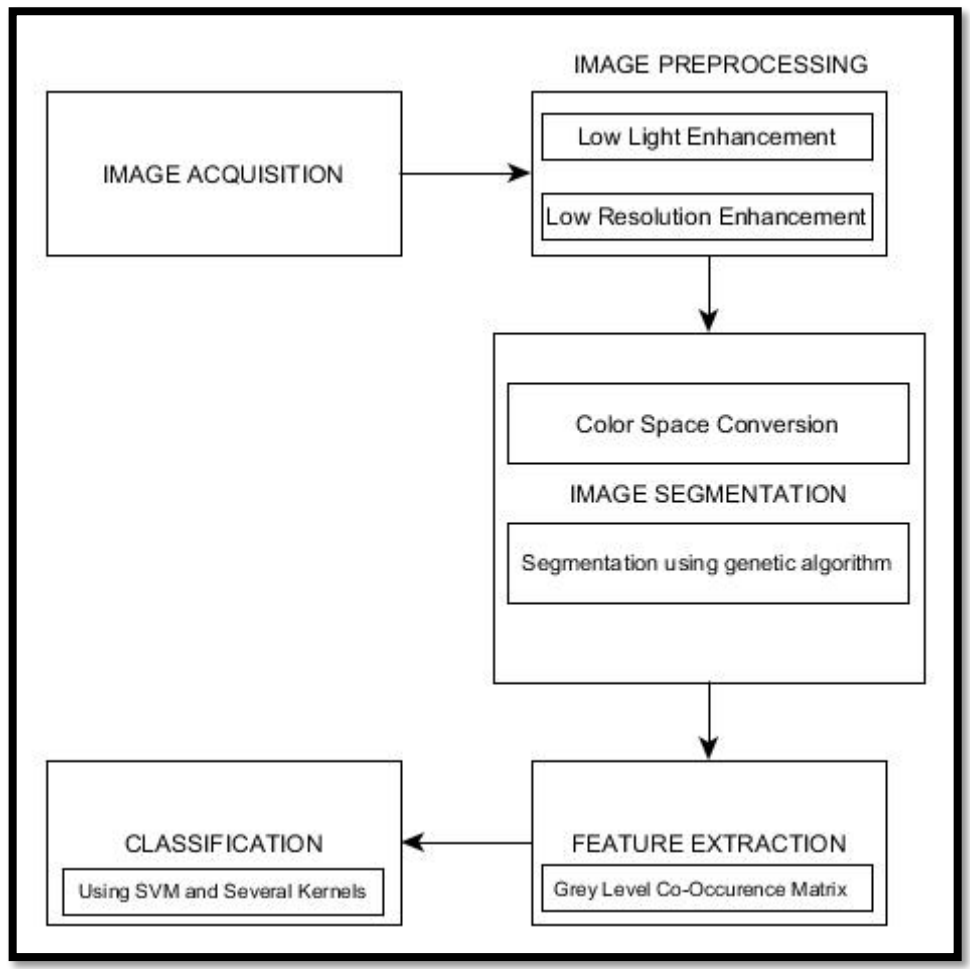

Figure.1 Architecture of Proposed Methodology

1. **Image Acquisition**

Images of the various leaves are captured and the input images are resized to a proper scale (for example 250x250 pixels) for performing the further operations. The image converted to digital format and are resized as to measure and compare different images.

**2. Image Preprocessing:**

The low light images and the low resolution images loose the details of the input images for the proper classification. To avoid this we use two methodologies in order to prevent these details. So, the low light image enhancement done by the Convolutional neural network using inception modules. And the low resolute images are improved by the adversarial neural networks.

## 2.1 Low Light Image enhancement:

Low light image enhancement belong to the low-level image processing technique. In this, we will be using the Convolutional neural network layers. In the Computer Vision, Deeper the network layers better the performance of the network with the high accuracy as expected. In this inception modules several layers are connected with the previous layers with the different format (1x1, 2x2, 3x3… 5x5). And the activation function used in this used are rectified linear unit function (ReLU). The output of the different layers are made into single resultant vector. The inception modules were used by the GoogleNet and have achieved high accuracy in classification for the challenge ImageNet Large-Scale Visual Recognition Challenge. We propose the network architecture with the special convolutional module as shown in Fig.2. This is divided into two stages. One way is by using the 3x3 convolutional modules and 1x1 convolutional layers and then we combine them together and input to the second stage. And the input data is directly bypassed with the shortcut easy route association used in the process of residual learning.

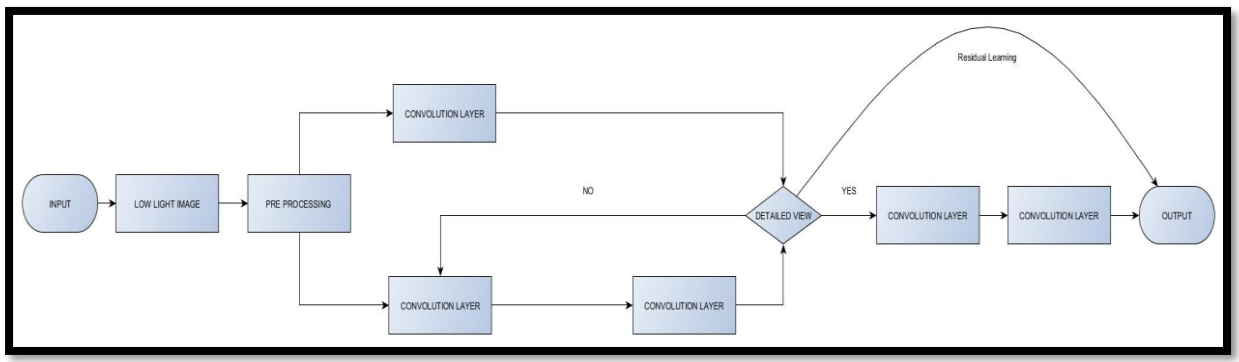

Figure. 2 Low Light enhancement architecture

Residual learning is proposed to use the overcome the problem with the deep networks. As the network goes deeper the network will face a serious issues and some of the issues are vanishing of the gradients. And this will obstruct the convergence during the training phase. In the ILSVRC15 challenge with the usage if the residual nets which has used 152 layers deep have been easily trained and has won the first place. This network will generate an image from the residual nets and the final resultant image is calculated by the residual image and the original image. Both the residual and the original image can be merged because the ground truth values doesn't differ much. And the Low light image enhancement is shown in Figure. 3.

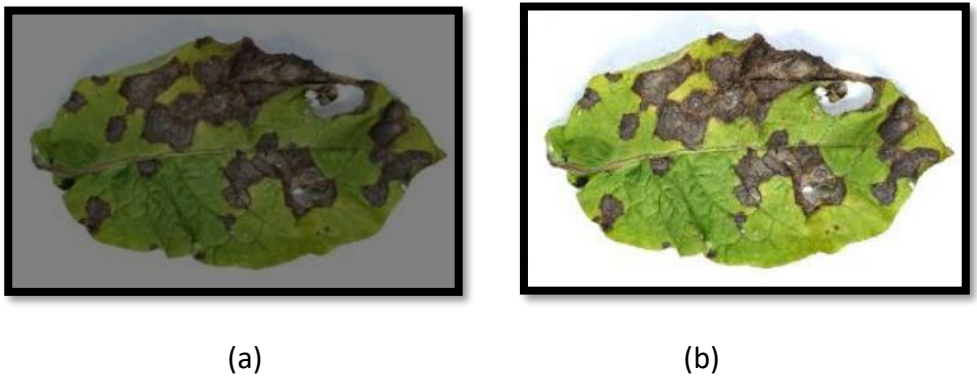

(a) (b)

**Figure. 3** The original image (a) which is low light image and after the enhancement (b)

### 2.1.1 Parameters for the Low Light Image Enhancement:

In the training data the low light images are given as input .Unlike the other image processing methods the ground truth values are not known and the low-light images are produced by the degradation method. The parameters for enhancing the images are gamma value ($\gamma$) and this randomly set between (1.5 to 5.5). This range will allow the network to improve the brightness of the original image adaptively. And the other parameter for our enhancement technique is SSIM loss. For, the low light image enhancement the texture preventing is the most important task and the intensity levels or brightness will fluctuate around the around value and the SSIM value lie around (0, 1). These parameters help the image to become brighter.

### 2.2 Low Resolution Image:

The low resolute images lose a lot of details in the images during the training phase and the images and this can be done by the Adversarial neural networks and this comes under the unsupervised learning and in this network which consists of the two networks and one network consist and the generative network and the other network is master network which will give the feedback to the generative network and the weights and bias of the network are adjusted in such a way that the texture of the network is preserved and the image pattern is carried out and image resolution even if it is low resolution the details of the image can be visualized properly after the image is generated through this adversarial generative network.

### 3. Image Segmentation:

### 3.1 Color Space Conversion:

In this session once the image brought to the suitable format then the data has to be converted to the different domain which makes the computing more convenient and more flexible and the observation of the RGB images is often striking different from the human perception and computes into proper representation. So, the color transformation is required to make the image more convenient form for the computation of computer. The original image which is in (RGB) color space which consists of the red component, blue component and green component are converted to the YC$_b$C$_r$ color space. This YC$_b$C$_r$ color space model representation is independent of device which represents color based on the human vision. The conversion of the image from RGB to YC$_r$C$_b$ can be done by using below transformation.

**Formula:**

$$\begin{bmatrix} Y \\ Cb \\ Cr \end{bmatrix} = \begin{bmatrix} 16 \\ 128 \\ 128 \end{bmatrix} + \begin{bmatrix} 0.2568 & 0.5041 & 0.0979 \\ -0.1482 & -0.2910 & 0.4392 \\ 0.4392 & -0.3678 & -0.0714 \end{bmatrix} \begin{bmatrix} R \\ G \\ B \end{bmatrix}$$

The Conversion of image from RGB to YC$_r$C$_b$ is shown in the figure .4.

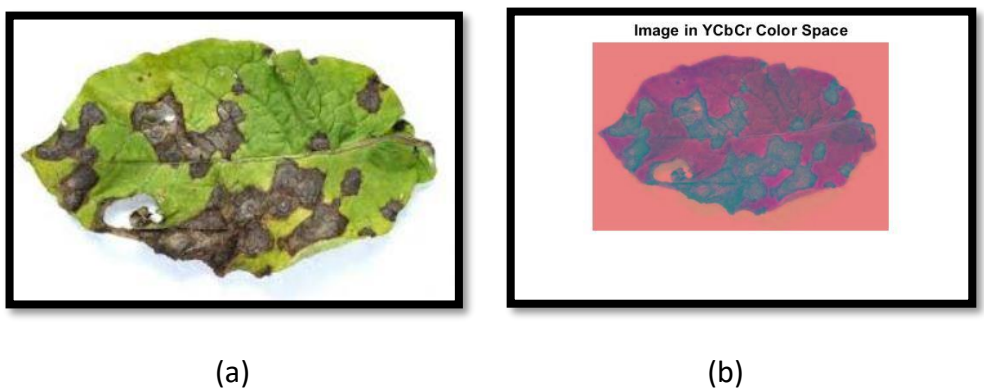

(a) (b)

**Figure. 4** The image (a) converting into the image in YC$_r$C$_b$ color space (b)

In the YC$_b$C$_r$ color space the Y is the luma component which represent the light intensity. C$_b$ and C$_r$ are the blue and red chroma components. Since the Y component is more sensitive than the human eye, it needs to be more correct and the C$_b$ and C$_r$ is less sensitive to the human eye. So, it uses this sensitive of the human eye remove the unnecessary details to the human eyes.

### 3.2 Segmentation using Genetic Algorithm Approach

Once the image brought to the convenient format and this session will be the representation of any image that gives more significance to the object of interest from the background and simpler to analyze based on the color using K- clusters. This session describes the highlights of the particular area of interest and then diving into groups. Here the value "K" represents the no of components has to be divided (here the each component are each color). And this segmentation into k clusters is done with the genetic algorithm. The process of segmentation is shown in the figure 5.

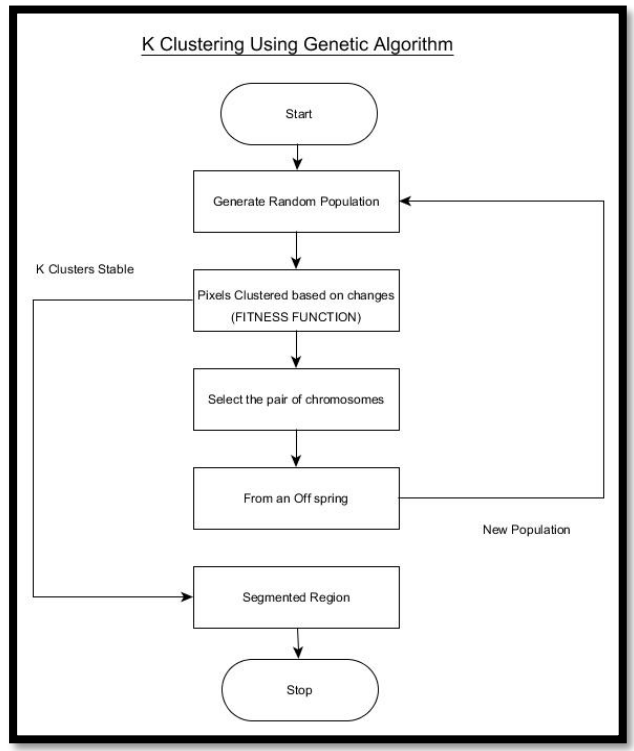

**Figure .5** Algorithm for clustering using Genetic Algorithm

The genetic algorithm is helpful in searching the large space such as instead of searching all the possible ways in the brute force the algorithm gets optimized in each and every phase. For finding the clusters we can use this large search capable genetic algorithm, And the set of points which are unlabeled can formed as k-clusters. Let's the color image of size m x n dimensions and every pixels has its respective color components. Every chromosomes has its corresponding solution which has the sequence of the k-cluster centers. The population is initialized in several rounds and already existing chromosomes the best chromosomes live or survive in every round and proceed for the further rounds processing.

The fitness function is one of the biggest concern. In the beginning of the computation of the fitness the pixels of the images are clustered according the distance between the randomly assigned cluster center and the pixel values. The nearest cluster center are chosen. The each pixel in the image is $x_i$ and these are adjusted into their respective cluster center and these can be represented as $Z_j$ for j= 1, 2, 3 ….K. where K is the no of clusters and by the following equation.

If    $||X_i-Z_j|| < ||X_i-Z_q||$,

Where i = 1, 2… m x n (All the pixel values of the Image)

      q=1, 2… k and (p is not equal to j)

After assigning the cluster centres randomly the mean (or) average of all the pixel values of the cluster are took and then the new cluster is updated. The new cluster centre of $Z_i$ is shown by clustering $c_i$ as:

$Z_i(y, cr, cb) = \sum_{xj \in ci} xj(y, cr, cb))$  where i = 1, 2 … k

Now, the fitness function is the own customized function which is figured by computing the Euclidian distance between the centre of the cluster and the pixels of the associated cluster and is shown by the following equations.

$M = \sum M_i$

$M_i = \sum_{xi \in ci} |xj(y, cr, cb) - zi(y, cr, cb)|$

And the clusters are formed based on the color and these are segmented based on the color. And the segmented regions based on the colors are shown in figure 6.

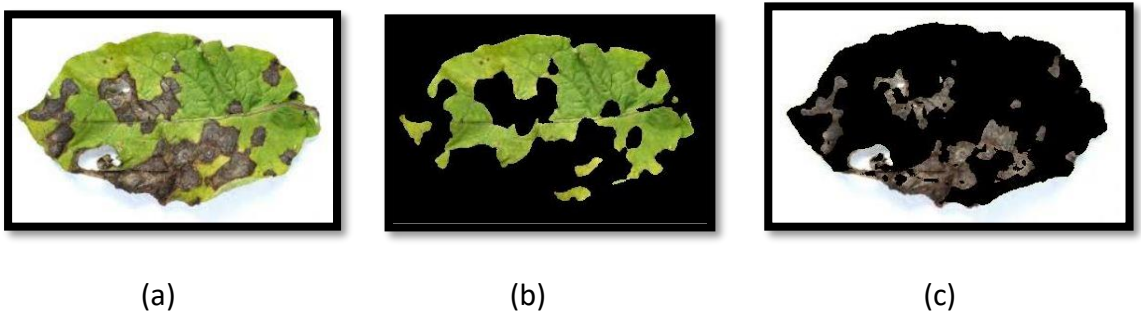

(a)          (b)          (c)

Figure. 6  An example of the sample input leaf image which is infected by anthracnose disease and the image are (a) enhanced imaged in the previous step (b) is the first cluster with mostly green pixels (c) the second cluster with infected region.

**4. Feature Extraction:**

After the segmentation of the area of interest which means the extraction of the diseased part from the image which is mostly the browned region in RGB format. There are many types of the feature extraction which usually include color, edge, shape, texture extraction. Targeting the plant leaf for the classifying the leaf surface texture is the key component for the classification. This phase aims for the feature extraction from the image for that purpose in this paper we exhibit a mathematical method using the GLCM. The GLCM stands for the Grey Level Co-occurrence Matrix. The GLCM is the statistical model for investigating the texture of the object. This describes the alignment of the pixels in the spatial relationship. The GLCM function is helpful for the describing the texture of any object by computing the co-occurrence matrix of the pixel value 'i' with respect to the pixel 'j' which are occurring in the spatial relationship. The matrix vectors of GLCM or the feature vectors for the sample leaf image is shown in Fig.7

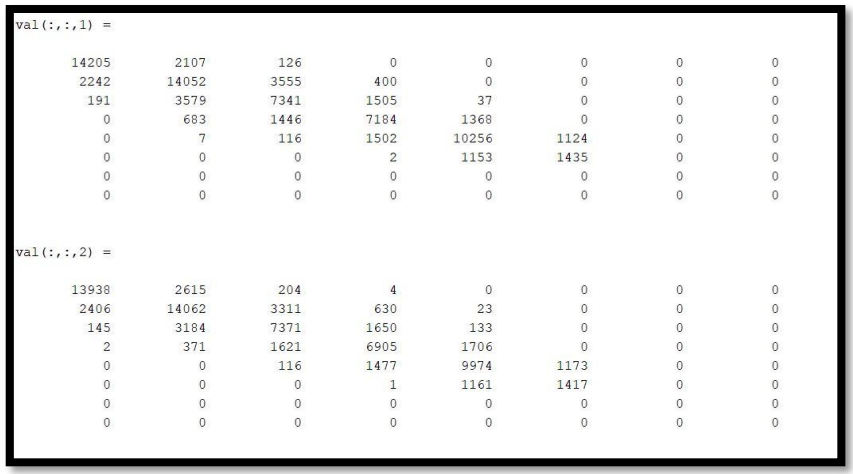

**Fig. 7** The feature vectors of the GLCM co-occurrence matrix.

1. Contrast: Contrast describes the intensity values of a pixel. Contrast of a pixel and its corresponding surrounded pixels in the entire image. This can be understood by assuming if the contrast value is 0 it implies that the whole image is constant or it will have high value if the neighbor pixel value varies with the high value.

$$\text{Contrast} = \sum_{i,j=0}^{N-1} P_{ij}(i-j)^2$$

2. Energy: Energy describes the uniformness in the image with square elements summation in the GLCM. This value generally varies between 0 and 1. If the value is 1 it means that the image is constant.

$$\text{Energy} = \sum_{i,j=0}^{N-1}(P_{ij})^2$$

3. Dissimilarity: While measuring the weights in the contrast measure as pixel goes away from the diagonal the weights increases drastically. The dissimilarity helps in measuring the weights linearly (0, 1, 2, 3,….).

$$\text{Dissimilarity} = \sum_{i,j=0}^{N-1} P_{ij}|i-j|$$

4. Entropy: Entropy is responsible for the information needed to compress the image. Entropy describes the amount of loss of data in a signal which are being transmitted and also measures the signal data.

$$\text{Entropy} = \sum_{i,j=0}^{N-1} P_{ij}\log P_{ij}$$

5. Correlation: Correlation describes how to correlate a pixel to its neighbour pixels in the entire image. And the values of the range in between (-1, 1).

$$\text{Correlation} = \sum_{i,j=0}^{N-1} P_{ij}(i-\mu)(j-\mu)/\sigma^2$$

Where $P_{i,j}$ the pixel value of the image at position i,j amd the N is the Number of gray levels.
Mean ($\mu$) = Mean value or Average of all pixel values in the relationship contributed by GLCM matrix.

$$\mu = \sum_{i,j=0}^{N-1} iP_{ij}$$

Variance ($\sigma^2$) = variance of the intensity values in the image.

$$\sigma^2 = \sum_{i,j=0}^{N-1} P_{ij}(i-\mu)^2$$

These parameters are helpful for analyzing the alignment of the image and the corresponding properties and the feature vectors are calculated with the Grey level co-occurrence matrix.

**Classifier:**

After the feature vectors are extracted then the feature values are given as input to the classifier and the classifier here used are support vector machine (SVM) with different kernels and the kernels used are Radial Basis Functional Kernel (RBF), Linear Kernel, Polygonal Kernel, and Quadratic Kernel. Here the purpose of the classifier is the different bacteria causes different diseases to the plant leaves and for the classification of the diseases of the plant based on the infected area we use this classification model and the output result will be of the classification of the each disease respectively such as Canker, Leaf spot, Anthracnose, and Blight. And for the classification, we have used a dataset of 235 images in which we have split the dataset into the training and the validation data and the training set consists of images and 215 images for the validation set and the accuracy is validated using the formulae.

$$\text{Accuracy (\%)} = \frac{No.\,of\ correctly\ classified}{Total\ no\ of\ leaves\ in\ the\ datset} \times 100$$

**IV. Results:**

All the modules of the experiment are conducted in the MATLAB software. The input images are the plant leaf images like Tomato leaf, beans leaf, pepper leaf, orchid plant leaf, and these leaves are affected by the bacteria. And these different bacteria cause different diseases to the plant leaves and they are classified and outputted from the experiment. Here the feature vectors are in the form of the co-occurrence matrix these are calculated after mapping the corresponding color component of the input image. The co-occurrence matrix (features) of the input image (leaf image) are computed and then these values are compared with the corresponding feature values present in the training set data features values. The classification is done and viewed with the MATLAB GUI using guide. And is shown in figure 8.

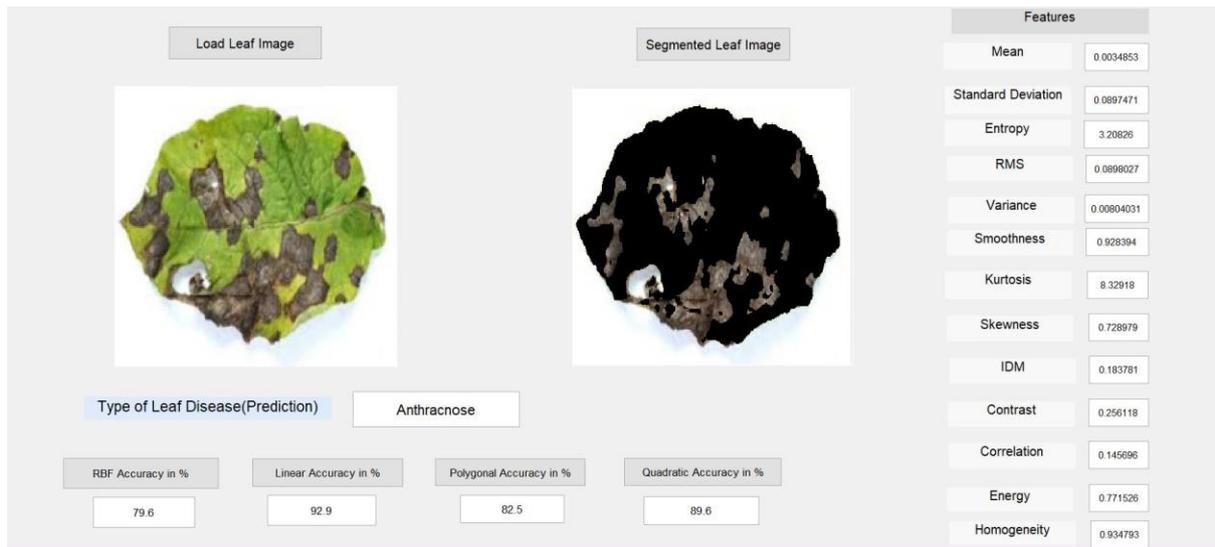

**Figure 8** Graphical User Interface in MATLAB for the classification of the leaf for the disease detection. And the corresponding accuracy for the kernels and the features on the right side of the image

And now for the testing purpose a complete new image which is not present in the training and the validation dataset are chosen. And the accuracy using the SVM with the different kernels are shown in the Table 2. And the Accuracy of each diseases leaf dataset in the validation set is shown in Table 1.

| Leaf Disease | No.of Validation Images | Correctly classified(✓) | Incorrectly classified(✗) | Accuracy (%) |
|---|---|---|---|---|
| Blight | 64 | 54 | 10 | 84.37% |
| Anthracnose | 56 | 51 | 5 | 91.07% |
| Canker | 58 | 50 | 8 | 86.20% |
| Leaf Spot | 37 | 26 | 11 | 71.02% |

**Table. 1** Result table

| Leaf Image | Label | RBF Kernel | Linear Kernel | Polygonal Kernel | Quadratic Kernel |
|---|---|---|---|---|---|
| 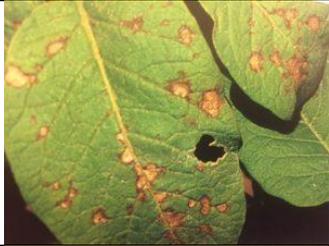 | Blight | 80.5% | 92.7% | 81.5% | 87.6% |
| 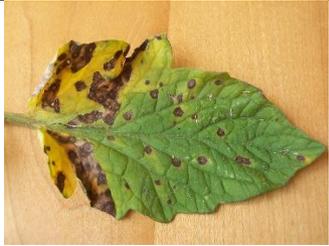 | Anthracnose | 78.9% | 91.5% | 85.1% | 82.9% |
| 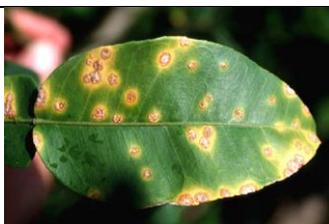 | Canker | 77.3% | 87.5% | 83.5% | 88.7% |
| 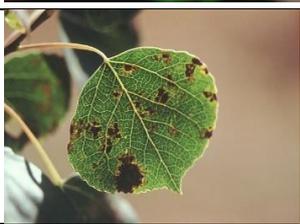 | Leaf Spot | 82.6% | 91.3% | 78.0% | 89.5% |

**Table .2** The accuracy of the new images and their corresponding accuracy with the different kernels.

**V. Conclusion and future work:**

This paper presents various image processing techniques for the detection of the infected part of the leaf. And this helps in identification of the disease in the early stages so there is a chance protecting the infected plants. In this paper we have proposed a method for low light image enhancement using CNN which uses the inception modules followed by residual learning and for the low resolution images we have used properties of generative adversarial networks. Upton this process makes the input image proper and for the better understanding extracting the area of interest we propose an algorithm for segmentation using genetic algorithm which is an optimization algorithm and requires very less computational power and very efficient. And for the classification of diseases we use an SVM. For the classification of the image we require feature vectors (texture as features) for this we statistical model called GLCM. Here in classification different kernels have been used and has achieved high accuracy in classification of diseases. In the further research in the classification or accuracy Artificial Neural Networks and many other deep learning models and naturally inspired optimization algorithm can also be used.

## VI. References

**[1]** Singh, V., & Misra, A. K. (2017). Detection of plant leaf diseases using image segmentation and soft computing techniques. Information Processing in Agriculture, 4(1), 41-49.

**[2]** Arya, M. S., Anjali, K., & Unni, D. (2018, January). Detection of unhealthy plant leaves using image processing and genetic algorithm with Arduino. In 2018 International Conference on Power, Signals, Control and Computation (EPSCICON) (pp. 1-5). IEEE.

**[3]** Kaur, L., & Laxmi, V. (2016). Detection of Unhealthy Region of plant leaves using Neural Network. Disease management, 1(05), 34-42.

**[4]** Madhogaria, S., Schikora, M., Koch, W., & Cremers, D. (2011). Pixel-based classification method for detecting unhealthy regions in leaf images. In Proc. of 6th. Workshop on Sensor Data Fusion (SDF).

**[5]** Choras, R. S. (2007). Image feature extraction techniques and their applications for CBIR and biometrics systems. International journal of biology and biomedical engineering, 1(1), 6-16.

**[6]** Sai, S. M., Naresh, K., RajKumar, S., Ganesh, M. S., Sai, L., & Nav, A. (2018, April). An Infrared Image Detecting System Model to Monitor Human with Weapon for Controlling Smuggling of Sandalwood Trees. In *2018 Second International Conference on Inventive Communication and Computational Technologies (ICICCT)* (pp. 962-968). IEEE.

**[7]** Gavhale, K. R., Gawande, U., & Hajari, K. O. (2014, April). Unhealthy region of citrus leaf detection using image processing techniques. In Convergence of Technology (I2CT), 2014 International Conference for (pp. 1-6). IEEE.

**[8]** Sarayloo, Z., & Asemani, D. (2015, May). Designing a classifier for automatic detection of fungal diseases in wheat plant: By pattern recognition techniques. In Electrical Engineering (ICEE), 2015 23rd Iranian Conference on (pp. 1193-1197). IEEE.

**[9]** Arya, M. S., Anjali, K., & Unni, D. (2018, January). Detection of unhealthy plant leaves using image processing and genetic algorithm with Arduino. In 2018 International Conference on Power, Signals, Control and Computation (EPSCICON) (pp. 1-5). IEEE.

**[10]** Shaikh, R. P., & Dhole, S. A. (2017, April). Citrus leaf unhealthy region detection by using image processing technique. In Electronics, Communication and Aerospace Technology (ICECA), 2017 International conference of (Vol. 1, pp. 420-423). IEEE.

**[11]** Singh, V., & Misra, A. K. (2015, March). Detection of unhealthy region of plant leaves using Image Processing and Genetic Algorithm. In Computer Engineering and Applications (ICACEA), 2015 International Conference on Advances in (pp. 1028-1032). IEEE.

**[12]** Sasi, K., Radhiga, P., & Chandraprabha, K. (2018). Plant Leaf Disease Prediction and Solution using Convolutional Neural Network Algorithm. International Journal, 6(4).

**[13]** Dhaygude, S. B., & Kumbhar, N. P. (2013). Agricultural plant leaf disease detection using image processing. International Journal of Advanced Research in Electrical, Electronics and Instrumentation Engineering, 2(1), 599-602.

**[14]** Kulkarni, A. H., & Patil, A. (2012). Applying image processing technique to detect plant diseases. International Journal of Modern Engineering Research, 2(5), 3661-3664.

**[15]** Bashir, S., & Sharma, N. (2012). Remote area plant disease detection using image processing. IOSR Journal of Electronics and Communication Engineering, 1(6), 31-34.

**[16]** Chaudhary, P., Chaudhari, A. K., Cheeran, A. N., & Godara, S. (2012). Color transform based approach for disease spot detection on plant leaf. International Journal of Computer Science and Telecommunications, 3(6), 65-70.